\pdfoutput=1

\documentclass[11pt]{article}

\usepackage[final]{acl}

\usepackage{times}
\usepackage{latexsym}

\usepackage[T1]{fontenc}

\usepackage[utf8]{inputenc}

\usepackage{microtype}

\usepackage{inconsolata}

\usepackage{graphicx}
\usepackage{tcolorbox}
\tcbuselibrary{skins}
\usepackage{amsmath}
\usepackage{amsfonts}
\usepackage[frozencache=true,cachedir=minted-cache]{minted}
\usepackage{booktabs}
\usepackage{fancyvrb}
\usepackage{multirow}
\usepackage{adjustbox}
\usepackage{pgfplots}  
\pgfplotsset{compat=1.17}
\usepackage{xspace}
\usepackage{float}
\usepackage{makecell}
\usepackage{hyperref}
\usepackage{etoolbox}
\usepackage{xcolor}
\usepackage{colortbl}

\newcommand{\fans}{{FANS}\xspace}

\usepackage{color}
\usepackage{listings}
\definecolor{keywordcolor}{rgb}{0.7, 0.1, 0.1}   
\definecolor{tacticcolor}{rgb}{0.0, 0.1, 0.6}    
\definecolor{commentcolor}{rgb}{0.4, 0.4, 0.4}   
\definecolor{symbolcolor}{rgb}{0.0, 0.1, 0.6}    
\definecolor{sortcolor}{rgb}{0.1, 0.5, 0.1}      
\definecolor{attributecolor}{rgb}{0.7, 0.1, 0.1} 

\newtoggle{anonymous}
\toggletrue{anonymous}
\newcounter{tcb}

\title{FANS: \underline{F}ormal \underline{An}swer \underline{S}election for LLM \\Natural Language Math Reasoning Using \emph{Lean4}}

\author{
 \textbf{Jiarui Yao\textsuperscript{1}},
 \textbf{Ruida Wang\textsuperscript{1}},
 \textbf{Tong Zhang\textsuperscript{1}}
\\
 \textsuperscript{1} University of Illinois Urbana-Champaign\\
 \texttt{
   \{jiarui14, ruidaw, tozhang\}@illinois.edu
 }
}

\begin{document}
\maketitle
\begin{abstract}
Large Language Models (LLMs) have displayed astonishing abilities in various tasks, especially in text generation, classification, question answering, etc.
However, the reasoning ability of LLMs still faces many debates, especially in math reasoning. The inherent ambiguity of Natural Language (NL) limits LLMs' ability to perform verifiable reasoning, making the answers lack coherence and trustworthy support.
To tackle the above challenges, we propose a novel framework named \textbf{FANS}: \textit{\underline{F}ormal \underline{AN}swer \underline{S}election for LLM Natural Language Math Reasoning Using Lean4}.
\iftoggle{anonymous}{}{
\footnote{Our code could be found at \url{https://github.com/MaxwellJryao/TheoremVerifier}.}
}
It is a pioneering framework that utilizes Lean4 to enhance LLMs' NL math reasoning ability.
In particular, given an NL math question and LLM-generated answers, \fans first translates it into Lean4 theorem statements. Then it invokes another Lean4 prover LLM to produce proofs, and finally verifies the proofs by Lean4 compiler. Answers are selected based on the verifications.
It enhances LLMs' NL math ability in providing a computer-verifiable solution for its correct answer and proposes an alternative method for answer selection beyond the reward model based ones.
Our experiments demonstrate the effectiveness of \fans with an improvement of nearly $2\%$ across several math benchmarks, and even higher further based on reward models or in subfields such as algebra and number theory that Lean4 is better at. The code is available in \href{https://github.com/MaxwellJryao/FANS}{https://github.com/MaxwellJryao/FANS}.
\end{abstract}
\section{Introduction}

\begin{figure}[t]
    \centering
    \includegraphics[width=\columnwidth]{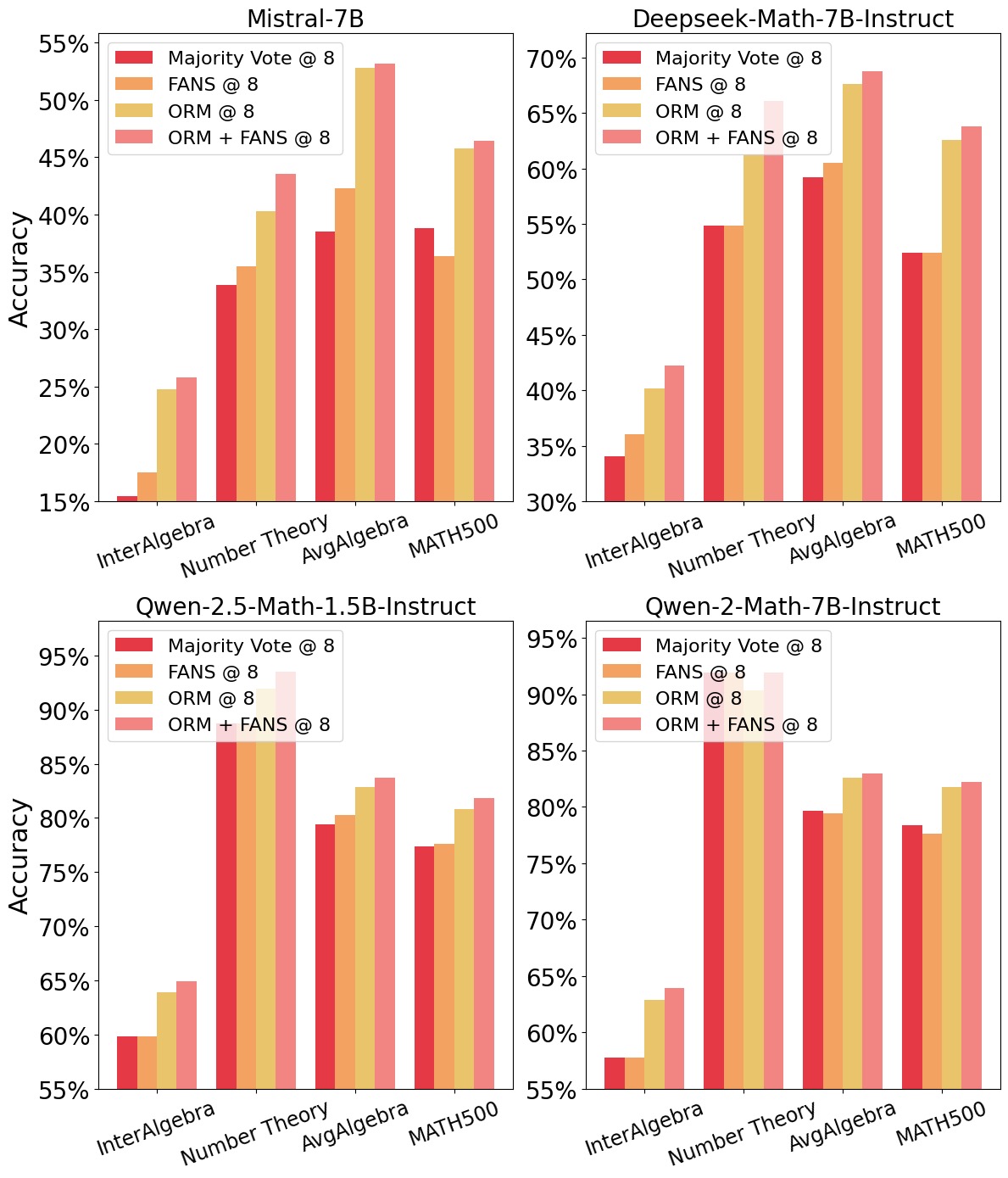}
    \caption{Comparision between \fans and majority vote, together with ORM and ORM + \fans method. From the results, we could see \fans based on ORM achieve the highest accuracies consistently across different base models and different test sets. In particular, \fans performs well on the sub-fields of number theory and algebra, which are better supported by Lean4 with its existing libraries.}
    \label{fig:main result}
\end{figure}

Math reasoning and problem-solving capability have been a hot topic in LLM's research field. Researchers are trying to develop new LLMs~\cite{shao2024deepseekmath, yang2024qwen25mathtechnicalreportmathematical, yu2023metamath}, increasingly harder benchmarks~\cite{cobbe2021training, hendrycks2021measuring}, and advanced reward model methods for evaluation of answers~\cite{dong2023raft, yang2024qwen25mathtechnicalreportmathematical}.
However, there have always been debates on whether LLMs are able to perform real reasoning or simply do pattern matching~\cite{saparov2023testing}. Consequently, researchers have begun to consider using symbolic languages to perform computer-verifiable formal reasoning in mathematics. Consequently, many formal mathematical languages have been proposed such as Lean~\cite{de2015lean, moura2021lean}, Isabelle~\cite{paulson1994isabelle}, and HOL Light~\cite{harrison2009hol}. These Formal Languages (FLs) map existing mathematics into their formal kernels, which allows computers to automatically verify mathematical proofs. FL provides a clear standard for evaluating answers to theorem proof and significantly impacts both mathematical and computer science communities.

\begin{figure*}[]
    \centering
    \includegraphics[width=\linewidth]{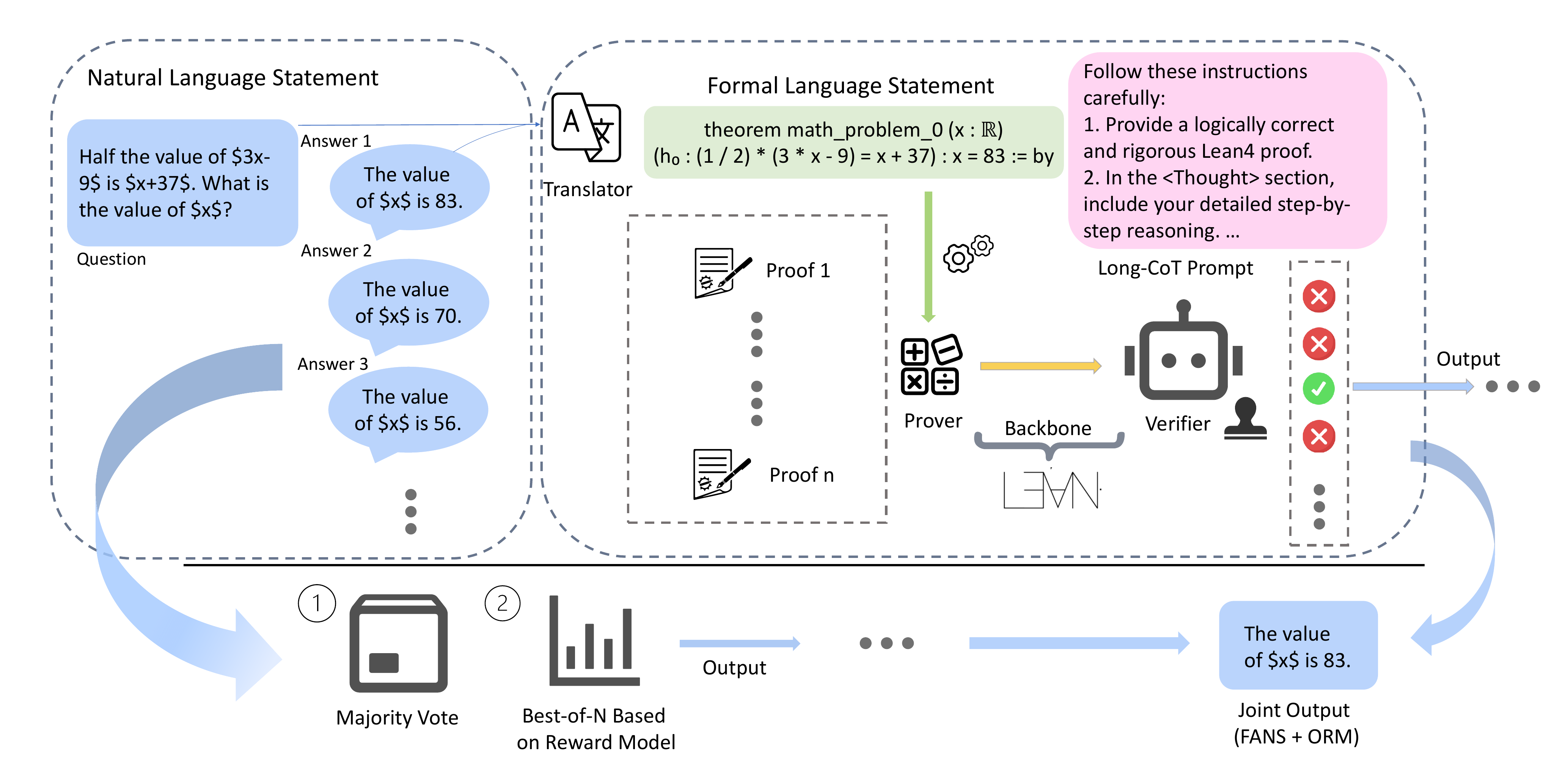}
    \caption{\textbf{FANS Framework:} The framework shown in the upper part first passes the Natural Language math questions and the LLM-generated answers to our Long CoT NL-to-FL translator. Subsequently, it invokes a prover to prove the translated Lean4 statements and uses the verifier to check whether the proofs are correct. The correct outputs are used for further answer selection as a verifiable foundation. \textbf{Existing Methods:} majority vote and best-of-N ranking based on reward models are shown in the lower part of the figure.}
    \label{fig:enter-label}
\end{figure*}

In the context of formal reasoning, most existing works focus on how to use NL to enhance the FL capability of the model~\cite{xin2024deepseek, wang2024theoremllama, lin2024lean, wang2025malot}. Research has also shown that formal language solutions to a math problem can be translated back to a valid natural language proof with very few errors~\cite{jiang2022draft}. Despite extensive research on formal languages (FLs), few have successfully leveraged Lean4 or other formal languages to enhance LLMs' natural language based mathematical reasoning. On the other hand, for complex natural language math reasoning problems, most LLMs struggle to find a correct solution in limited attempts, but could eventually provide the correct solution through multiple attempts. This leads to a natural idea of scaling the test-time computation resources in NL reasoning to get the correct answer. ~\citet{dong2024rlhf,shao2024deepseekmath, yang2024qwen25mathtechnicalreportmathematical} shown that a model with scale around 7B without advanced fine-tuning methods could achieve a pass@n accuracy around $90\%$ on MATH500~\cite{lightman2023lets} after a sufficient number of trials. Much recent literature has revealed the effectiveness of test-time scaling for improving the model's performance~\cite{muennighoff2025s1, chen2024expanding, snell2024scaling, wu2024inference, guan2025rstar}. 

However, this raises an important question of how to select the correct answer from outputs from the model, which is highly non-trivial especially in the case that we do not have the ground truth answer. Fundamentally speaking, this partially originates from mathematical reasoning being a task with a solid logical foundation, but natural language's inherent ambiguity makes the generated results not trustworthy enough. To solve this problem, many answer-selection methods have been proposed, ranging from majority vote to best-of-N ranking built upon various reward models~\cite{dong2024rlhf, yang2024qwen25mathtechnicalreportmathematical}. But these methods are either too simple to fulfill the pass@n potential, or still pure LLM-based, failing to make use of the solid formal foundation of math reasoning. In addition, it is hard for reward models to generalize to out-of-distribution (OOD) domains or models due to the distribution shift.

To solve the above challenges, we propose our novel \fans framework. It contains the following stages that work synergistically to enhance answer selection and provide a verifiable formal foundation for NL math reasoning:

\noindent \textbf{(a) NL to FL translation:} Based on the outstanding reasoning of Long Chain-of-Thought (CoT), we propose the training method for the Long CoT translator named \textit{LeanTranslator}. It can translate general math question-answering problems into Lean4 provable formal statements. Applying the translator to question-answer pairs generated by LLMs in NL math reasoning, we can obtain Lean4 statements for further processing.

\noindent \textbf{(b) Lean4 proof writing and verification:} We pass the generated statement to an advanced prover to generate the proofs for the FL statement, then use the Lean4 verifier to judge the correctness of the proofs. Final answer selection will be based on the verification results.
\noindent \textbf{(c) Answer selection and output:} After we obtain the verification results, we can combine them with existing answer selection methods jointly. Since our answers are formally provable, it conveys a more solid foundation and trustworthiness than direct NL reasoning.

We summarize our contributions as follows: 
\begin{enumerate}
    \item We propose \fans, a pioneering framework that applies formal mathematical language, Lean4, in enhancing the capability of LLMs to solve NL math problems. 
    \item In our framework, we propose a new method to train the Long CoT translator that translates NL question-answer pairs into Lean4 statements. To be best of our knowledge, this is the first work that proposes such methods.
    \item Extensive experiments indicate that enhanced by our framework, LLMs are able to perform better answer selection. It can improve the accuracy on MATH500 by at most \textbf{1.91\%} and AMC23 by at most \textbf{8.33\%}. In some particular subfields that Lean is better at, we can even select \textbf{all} correct solutions. The qualitative studies also indicate that our framework can make the NL solutions generated by the model have a solid basis using formal language, enhancing the trustworthiness of the answer.
\end{enumerate}

\section{Methodology}

High-level speaking, \fans framework can be decomposed into three stages. Firstly, we introduce methods for translation for Natural Language (NL) question-answer pairs into their corresponding Formal Language (FL) statements in Section~\ref{meth:FLTranslate}. Subsequently, we detail how we write FL proofs and verify it using the Lean4 verifier in Section~\ref{meth:FLProof}. Finally, we introduce further usage for the verified proofs in Section~\ref{meth:further}. The general idea for \fans is to bridge the gap between FL and NL using the verified FL proofs as a solid foundation for LLM's NL reasoning.

\subsection{From Natural Language to Formal Language}\label{meth:FLTranslate}
To obtain accurately aligned FL statements from the original NL question-answer pairs, we first train a Long Chain-of-Thought (Long-CoT) \cite{openai2024reasoning} translator using pair-wise NL-FL statements. We try to teach a prover to analyze the NL statement in its Long CoT and then translate it into the FL statement. However, since there is no Long CoT data available for NL-to-FL statement translation, inspired by transfer learning methods for Lean prove Long CoT in MA-LoT~\cite{wang2025malot}, we introduce the following training method for FL statement translator training.



Firstly, we collect the NL-FL aligned statement translation data from Lean-Workbook~\cite{ying2024lean}. Based on such data, as versatile examples and use analysis-then-generate method inspired by \citet{wang2023let}, we generate NL statements for the DeepSeek-Prover-v1 dataset's Lean4 theorem statement using Qwen-2.5-72B. Altogether, we obtain a dataset containing 162,181 NL-FL aligned theorem statements without Long CoT. Among them, 21,967 are generated from DeepSeek-v1 and 140,214 are from Lean-Workbook.

Based on the aligned dataset, we train a Long CoT translator using the LoT-Solver~\cite{wang2025malot} as the base model. Since there is no Long CoT annotated theorem statement translation dataset, we apply the transfer learning technique. During training, we explicitly instruct the model in the system prompt to answer without using Long CoT and provide the empty Long CoT content. During inference, we instruct the model to use Long CoT to answer the question. In the inference, the model learns from the basic Long CoT ability in LoT-Solver to accurately translate the NL statement into the FL statement. Details of training and inference data examples can be found in Appendix~\ref{appendix:translatorExample}.

Besides our own Long CoT translator, recently, \citet{wang2025kimina} also released an NL-FL translator, and we also utilize it as the translator in the following experiments. Since it is not guaranteed that the translated FL statements are perfectly consistent with the original NL statements, we introduce another procedure to check whether the translations are faithful by prompting the LLMs themselves or a stronger one, QwQ-32B \citep{qwq32b} in our case, to generate a decision.

\subsection{Formal Language Proofs and Verification}\label{meth:FLProof}
After receiving the translated formal language statements, we utilize the open-sourced Lean4 provers \citep{lin2024Goedelprover,ren2025deepseek} to produce potential proofs written in Lean4 as well. To adapt the prover on the formal language statements from natural language math problems, we use a few-shot prompt for better alignment with the formats of classic formal language proof problems. Since the natural language statements have been translated into the standard formal language statements format, provers fine-tuned on formal language statements-proofs could be immediately applied without any modification. For this very reason, \fans could generalize among different backbone provers easily, with a highly disentangled formulation of its three-stage procedure.

The core backbone for \fans to be rigorously verified stems from the formal proof process in the functional programming level, which could be implemented by retrieving the compilation results from the verifier. At its core, Lean4 operates within the calculus of inductive constructions, ensuring that every proof is mechanically verified through a type-checking system that enforces strict logical consistency. This could eliminate human errors commonly found in formal reasoning, getting rid of unstable reasoning based on intuition and vague derivation. Therefore, once a theorem is proven in Lean4, it is mathematically indisputable, providing a robust foundation for tasks such as answer selection in math problem solving.  

\begin{table*}[htb]
    \centering
    \begin{adjustbox}{max width=0.9\linewidth}
    \begin{tabular}{lcccc}
        \toprule\midrule
        \textbf{Models}   & \textbf{MATH}(total)      & \textbf{MATH}-Algebra     & \textbf{MATH}-Number Theory   & \textbf{AMC23} \\
        \midrule
        \textbf{Mistral-MV}                 & 33.80             & 42.74             & 33.87             & 12.50 \\
        \textbf{Mistral-FANS}               & 36.40             & 45.97             & 35.48             & 15.00\\
        \textit{Improvement(\%)}            & \textbf{+7.69}    & \textbf{+7.56}    & \textbf{+4.75}    & \textbf{+20.00}\\
        
        \midrule
        \textbf{DeepSeek-Math-ORM}          & 62.60             & 82.26             & 61.29             & 30.00 \\
        \textbf{DeepSeek-Math-ORM+FANS}     & 63.80             & 82.26             & 66.13             & 32.50 \\
        \textit{Improvement(\%)}            & \textbf{+1.91}    & -                 & \textbf{+7.90}    & \textbf{+8.33}  \\    
        \midrule
        \textbf{Qwen-2.5-Math-ORM}          & 80.80             & 96.77             & 91.94             & 70.00 \\
        \textbf{Qwen-2.5-MATH-ORM+FANS}     & \textbf{81.80}    & \textbf{98.39}    & \textbf{93.55}    & 70.00 \\
        \textit{Improvement(\%)}            & \textbf{+1.25}    & \textbf{+1.67}    & \textbf{+1.75}    & -  \\ 
        \bottomrule
    \end{tabular}
    \end{adjustbox}
    \caption{Accuracies on math benchmarks (\%). For all baselines, we apply ORM@8 for comparison to using \fans based on ORM or using \fans compared with majority voting. The result for \fans is denoted as FANS-model.}
    \label{tab:main}
\end{table*}

\begin{table*}
    \centering
    \begin{adjustbox}{max width=0.95\linewidth}
        \begin{tabular}{lccccc}
        \toprule\midrule
        Models & MATH500 & Minerva Math & Olympiad Bench & AMC23 & Average \\
        \midrule
        Llama-3.2-3B-Instruct & 56.60 & \textbf{24.26} & 21.93 & \textbf{35.00} & 34.45\\
        FANS w/ self check & 57.60 & 22.06 & 21.89 & 32.50 & 33.51 \\
        \rowcolor{cyan!20} FANS w/ external check & \textbf{61.00} & \textbf{24.26} & \textbf{24.15} & \textbf{35.00} & \textbf{36.10} \\
        \rowcolor{gray!20}{FANS remove} & 61.80 & 27.21 & 25.48 & 40.00 & 38.62\\
        \midrule
        Deepseek-Math-7B-Instruct & 54.00 & \textbf{27.57} & 20.74 & \textbf{32.50} & 33.70\\
        FANS w/ self check & 55.40 & 26.47 & 22.37 & 27.50 & 32.94\\
        \rowcolor{cyan!20} FANS w/ external check & \textbf{57.00} & 27.21 & \textbf{22.67} & 30.00 & \textbf{34.22} \\
        \rowcolor{gray!20} FANS remove & 58.60 & 30.88 & 23.85 & 32.50 & 36.46\\
        \midrule
        Qwen2.5-Math-1.5B-Instruct & 79.00 & 33.46 & 44.30 & 55.00 & 52.94\\
        FANS w/ self check & 79.20 & \textbf{33.82} & 44.44 & 55.00 & 53.12\\
        \rowcolor{cyan!20} FANS w/ external check & \textbf{80.00} & \textbf{33.82} & \textbf{44.74} & \textbf{57.50} & \textbf{54.02}\\
        \rowcolor{gray!20} FANS remove & 80.20 & 36.40 & 45.63 & 60.00 & 55.56\\
        \midrule
        Qwen2.5-Math-7B-Instruct & 87.40 & \textbf{41.18} & 50.22 & \textbf{72.50} & 62.83\\
        FANS w/ self check & 87.40 & 40.07 & 50.52 & \textbf{72.50} & 62.62\\
        \rowcolor{cyan!20} FANS w/ external check & \textbf{88.00} & \textbf{41.18} & \textbf{50.67} & \textbf{72.50} & \textbf{63.09}\\
        \rowcolor{gray!20} FANS remove & 88.80 & 42.65 & 51.85 & 72.50 & 63.95\\
        \bottomrule
\end{tabular}
    \end{adjustbox}
    \caption{Results (\%) with new translator \citep{wang2025kimina} and current SOTA prover \citep{ren2025deepseek}.}
    \label{tab:res kimina-trans deepseek-new-prove}
\end{table*}


\subsection{Beyond Formal Language Itself}\label{meth:further}
Though the provers could achieve a $60\%$ to $80\%$ success rate on standard formal language proofs datasets like miniF2F \cite{zheng2021minif2f}, there are still chances that all the tries of formal proofs fail because of the gap between the formal language problems used to train the prover and the formats of translated formal language statements from the natural math QA questions. In this case, \fans will fall back to other alternate methods to select the best output.
For example, we could turn to the vanilla majority vote, or take advantage of reward models and output the answer with the highest score. For the former, we first conduct a majority vote on those answers successfully verified by the Lean4 verifier, and if there the number of votes for the winner is below one pre-specified threshold, we discard the answer and return to the majority vote directly to mitigate the impact of false positives due to mis-translation.

If the reward models are accessible, they could serve as a metric to determine which problems are more difficult by comparing the scores on specific questions and model outputs. Intuitively, harder problems might lead to more erroneous attempts, rendering the best-of-N based on reward models ineffective. Under this scenario, we resort to verifiable formal language proofs to accurately identify the correct answer among multiple wrong options.
\section{Experiments}\label{sec:exp}

We conduct extensive experiments on the MATH500~\cite{hendrycks2021measuring} and AMC23~\cite{yang2024qwen25mathtechnicalreportmathematical} to evaluate the effectiveness of \fans on enhancing the NL mathematical reasoning using FL methods. We quantitatively measure its usage in answer selection in Section~\ref{exp:result} and qualitatively show its capability in enhancing the trustworthiness of NL math reasoning by providing a formal backbone in Section~\ref{exp:qualitative}. Additionally, we conduct a thorough ablation study to validate the importance of each module of our proposed framework.

\subsection{Experiment Setup}\label{exp:setup}

\subsubsection{Dataset and Task}\label{setup:data}

In this work, we evaluate \fans's capability in enhancing LLMs' NL reasoning on several challenging datasets. The first is MATH500~\cite{hendrycks2021measuring}, a dataset containing 500 high-school-level math problems ranging in 7 major fields, including precalculus, algebra, number theory, etc. 
Another dataset we use is AMC23 from the repo of Qwen-2.5-Math~\cite{yang2024qwen25mathtechnicalreportmathematical}. This dataset contains 40 high-school-level math competition questions, ranging from similar fields as MATH500 but harder and more versatile in the form of question types. Other datasets include Minverva Math \citep{lewkowycz2022solving} and Olympiad Bench \citep{he2024olympiadbench}.

The goal of \fans is to use the FL method to enhance the existing answer selection methods like Majority Vote (MV) and Optimized Reward Model (ORM), and use formal reasoning to provide verifiable support for NL reasoning.

\subsubsection{Baselines}\label{setup:baseline}

To demonstrate how \fans works across diverse base models with different levels of answering mathematical questions, we select Mistral-7B~\cite{jiang2023mistral}, DeepSeek-Math-7B~\cite{shao2024deepseekmath} and Qwen-2.5-Math-1.5B / 7B~\cite{yang2024qwen25mathtechnicalreportmathematical}. The former two are relatively weaker models, leaving more space for them to be improved by \fans itself. In contrast, for the latter stronger models, which could answer more problems correctly, \fans majorly focus on providing trustworthiness and a verifiable foundation to the generated solutions. For ORM methods, we uniformly select Qwen-RM-72B because it is a larger model and has the best performance.

\subsection{Implementation Details}\label{exp:implementation}

In \fans, we use LoT-Solver as the base model for our Long CoT translator training using the transfer learning method we proposed. We train the translator on 162,181 records of NL-FL aligned statement data. To stabilize training, we also use block training and curriculum data sorting techniques in~\citet{wang2024theoremllama}. Besides, Kimina-Autoformalizer \cite{wang2025kimina} is also tested as the translator. We use DeepSeek-Prover-v1.5~\cite{xin2024deepseek}, Goedel Prover~\cite{lin2024Goedelprover} and DeepSeek-Prover-v2~\cite{ren2025deepseek} as our provers and \citet{santos2025kiminaleanservertechnical} as the verifier, which significantly reduces the verification overhead. 
We use NVIDIA H200 for model training and inference. Lean4 verification is conducted on CPUs.



\subsection{Results}\label{exp:result}

Table \ref{tab:main} demonstrates that \fans consistently improves the answer selection accuracy across all base models when compared to the baselines. On the two sub-fields, algebra and number theory, where lean4 performs better compared to other fields due to well-developed support from the language libraries, \fans could achieve an accuracy gain up to $7.90\%$. On stronger base models like Qwen-2.5-Math \cite{yang2024qwen25mathtechnicalreportmathematical}, \fans increases the accuracy for answer selection by $1.75\%$ as well. On the harder dataset AMC23, \fans helps weaker models to select the correct answer successfully, confirming that \fans working upon the verifiable proof process would be more helpful to identify the right answer from multiple wrong ones.

Table \ref{tab:res kimina-trans deepseek-new-prove} displays the results with the translator from \citet{wang2025kimina} and the current SOTA prover from \citet{ren2025deepseek} at 7B scale. Here ``FANS remove'' means we remove all false positive translations by eliminating those items with incorrectly selected answers but successfully verified FL proofs. Therefore, it could be regarded as a kind of upper bound for \fans, while an exact one since, if the translations are correct, provers could prove more FL statements. To ensure better consistency and fidelity of translated FL statements with the original NL statements, we introduce an extra stage to check whether the translations are correct or not, by invoking the base models themselves, or an external stronger model, for example, QwQ-32B \citep{qwq32b} here.

After the verification of the translations from the external model, \fans could achieve uniformly better final performance compared to all baselines, while the verification from the same base models could also achieve slight improvements on some datasets. This implies the potential of integrating FL to assist the answer selection in LLMs' math reasoning, especially by providing a verifiable backbone when the translation from natural language to formal language is faithful enough.

\subsection{Ablation Studies}\label{exp:abl}

\subsubsection{Dropping Long CoT in Translator}\label{abl:CoT}

To validate the effectiveness of the Long CoT translator in \fans, we conduct the study on dropping the Long CoT translator and replacing it with other models. Choices include the fine-tuned translator without Long CoT and GPT-4o-mini. We use few-shot prompts for all models. The results are presented in Table~\ref{tab:compare translator}. The results show that the accuracy for MATH500 without the Long CoT translator drops significantly, confirming the effectiveness of Long CoT in providing a more faithful translation. Comparison to the results on GPT-4o-mini shows that existing closed-source LLMs have suboptimal performance on FL translation, aligned with previous studies~\cite{wang2024theoremllama}.

\subsubsection{Using different reward models}\label{abl:RM}
To test the effectiveness of reward models in \fans, we conduct the ablation study on switching between different kinds of Reward models. In the experiment, we use different reward models to select data generated by Mistral-7B. The results are presented in Table~\ref{tab:compare rm}. It indicates that the Mistral reward model has the best performance on Mistral generated data while other reward models all suffer from suboptimal performances. This experiment shows that reward models may be unable to generalize to OOD model or data. Indicating the need for our generalizable answer selection methods.


\subsection{Qualitative Studies}\label{exp:qualitative}




\begin{table}[]
    \centering
    \begin{adjustbox}{max width=\linewidth}
    \begin{tabular}{lccc}
    \toprule\midrule
        Dataset         & \makecell{FANS \\(GPT-4o-mini)}   & \makecell{FANS\\(w/o Long CoT)}   & \makecell{FANS} \\
        \midrule
        Number Theory   & 30.65                             & \textbf{37.10}                  & \textbf{37.10}\\
        Prealgebra      & 57.32                             & 59.76                             & \textbf{60.98} \\
        Inter Algebra   & 16.49                             & 14.43                             & \textbf{17.53} \\
        Algebra         & 46.77                             & 48.39                             & \textbf{50.81} \\
        Precalculus     & 14.29                             & \textbf{21.43}                    & 19.64\\
        MATH500-Full    & 33.80                             & 35.40                             & \textbf{37.40} \\ 
    \bottomrule
    \end{tabular}
    \end{adjustbox}
    \caption{Comparison among different translators with Mistral-7B as the base model.}
    \label{tab:compare translator}
\end{table}

\begin{table}[t]
    \centering
    \begin{adjustbox}{max width=\columnwidth}\begin{tabular}{lcccc}
    \toprule\midrule
        Dataset & \makecell{ORM \\(Mistral)} & \makecell{ORM \\(Deepseek)} & \makecell{ORM \\(Qwen2.5\\-Math)} & \makecell{ORM \\(Qwen2.5\\-Math)\\ + FANS} \\
        \midrule
        Precalculus         & \textbf{30.36}     & 23.21     & 26.79     & 26.79 \\
        Prealgebra          & 64.63     & 65.84     & \textbf{73.17}     & \textbf{73.17} \\
        Interalgebra        & \textbf{25.77}     & 16.49     & 24.74     & \textbf{25.77} \\
        Algebra             & 56.45     & 52.42     & \textbf{60.48}     & \textbf{60.48} \\
        Number Theory       & 32.26     & 30.65     & 40.32     & \textbf{43.55} \\
        MATH500             & 42.40     & 38.80     & 45.80     & \textbf{46.40} \\
    \bottomrule
    \end{tabular}
    \end{adjustbox}
    \caption{Comparison among different reward models with Mistral-7B as the base model.}
    \label{tab:compare rm}
\end{table}

A concrete example through \fans pipeline is displayed in \ref{example of fans pipeline}. From the example, we could see that the translator translates the NL statement into its corresponding FL statement correctly, maintaining the original semantic meaning and mathematical formulation without inappropriate modification. In the proof stage, the prover successfully proves the FL statement with detailed step-by-step explanations preceding each line, which are omitted here for brevity. In the example, the question asks for the solution to a system of equations. Since the original natural language statement is already in a quite standard math format, our translator simply transforms the natural language description to a formal language expression. The prover generates a rigorous proof for the statement, which is successfully verified by the verifier. 

\section{Related Work}
\begin{figure*}[t]
\centering
\begin{adjustbox}{max width=0.9\linewidth}
\begin{tcolorbox}[title={Demonstration of \fans}]
\textbf{Natural Language Math Problem}

Find the product $CD$ of the integers $C$ and $D$ for which 
\(
\frac{C}{x-3}+\frac{D}{x+8}
=\frac{4x-23}{x^2+5x-24}\)
for all real values of $x$ except $-8$ and $3$.\\
 \textbf{Answer} $-5$.\\
 \textbf{Translated Formal Language Statement}
 \begin{lstlisting}
 theorem math_problem_0 (C D : ℤ) (h : ∀ x : ℝ, x ≠ 3 ∧ x ≠ -8 → C / (x - 3) + D / (x + 8) = (4 * x - 23) / (x ^ 2 + 5 * x - 24)) : C * D = -5 := by
 \end{lstlisting}
 \vspace{-1.3em}
 \textbf{Formal Language Proof}
 \begin{lstlisting}
theorem math_problem_0 (C D : ℤ) (h : ∀ x : ℝ, x ≠ 3 ∧ x ≠ -8 → C / (x - 3) + D / (x + 8) = (4 * x - 23) / (x ^ 2 + 5 * x - 24)) : C * D = -5 := by
    have h₁ := h 0 (by norm_num)
    have h₂ := h 1 (by norm_num)
    have h₃ := h (-1) (by norm_num)
    field_simp at h₁ h₂ h₃
    ring_nf at h₁ h₂ h₃
    norm_cast at h₁ h₂ h₃
    nlinarith
 \end{lstlisting}
 \vspace{-1.5em}
\end{tcolorbox}
\end{adjustbox}
\refstepcounter{tcb}
\label{example of fans pipeline}
\end{figure*}

Reasoning models, both proprietary like OpenAI-O1, Google Gemini Flash Thinking, Kimi-k1.5 \cite{MoonshotAI} and open-sourced ones such as Qwen Math \cite{yang2024qwen25mathtechnicalreportmathematical} and Deepseek-R1 \cite{xin2024deepseek1}, begin to substitute general language models in the core of LLM research. This stems from not only the representative of (math) reasoning ability for evaluating the intelligence of LLMs but also the various downstream tasks it could be applied to and the potential it reveals about the underlying immense abilities of LLMs to solve complicated problems that can only be resolved by humans in the past. \citet{zhou2024don} proposed a similar framework to \fans, but their method is heavily prompt-based, built upon weaker base models, while we train another Long CoT translator and utilize more advanced models to generate proofs and translation verification.

\subsection{Formal Language Reasoning}

The Formal Languages (FL) for math reasoning express mathematical statements in verifiable first-order logic. By solving math problems using FL, we can not only verify the correctness of the problem by the final answer like MATH~\cite{hendrycks2021measuring} or GSM8k~\cite{cobbe2021training} but also explicitly verify the correctness of each intermediate steps, making math reasoning has a solid foundation. Typical FL are like Isabelle~\cite{paulson1994isabelle}, Coq~\cite{coq1996coq}, Metamath~\cite{megill2019metamath}, HOL Light~\cite{harrison2009hol}, and Lean~\cite{de2015lean, moura2021lean}. Following ~\citet{yang2024leandojo}, we choose Lean4, the latest and most actively studied FL, as the language we use.

Traditional studies on FL all focused on how to annotate more data to boost the performance of LLMs on FL. Representative works like LeanDojo~\cite{yang2024leandojo} use retrial methods to select tactics; \citet{wang2024theoremllama} tries to use LLM to translate NL proof to FL proof; MA-LoT~\cite{wang2025malot} proposed Long CoT and multi-agent framework to solve FL questions; DeepSeek-Prover~\cite{xin2024deepseek, xin2024deepseek1}, Godel-Prover~\cite{lin2024Goedelprover}, and InternLM-Step-Prover~\cite{wu2024internlm2} applies massive data annotation to provide better foundation models. However, all of the above works focus on solving FL problems and ignore the potential of using FL to enhance the performance of NL math reasoning.


\subsection{Natural Language Math Reasoning}

Recent efforts to enhance the mathematical reasoning capabilities of large language models (LLMs) have spurred the development of diverse math problem-solving techniques.
Many works in this field focus on developing advanced foundation models for solving math word problems such as DeepSeek-Math~\cite{shao2024deepseekmath}, Qwen-2.5-Math~\cite{yang2024qwen25mathtechnicalreportmathematical}, Mistral-Math~\cite{yu2023metamath}, and Llemma~\cite{azerbayev2023llemma}. Other focus on the inference methods to query the existing models to write better answers, typical methods like traditional Chain-of-Thought (CoT)~\cite{wei2022chain,yao2025optimizing,xiong2025minimalist,zhang2025online}, tree-search methods~\cite{yao2023tree, jiang2024technical} and recent proposed Long-CoT approaches~\cite{openai2024reasoning, wang2025malot, open_o1}. However, none of them jump beyond the context of natural language in math reasoning, leading to LLMs' math reasoning without formal foundations.

\subsection{Reward Modeling} 

Reward models play a crucial role in assessing the quality of LLMs' outputs, serving as an essential component in downstream applications such as reinforcement learning from human feedback (RLHF). These models function as evaluative mechanisms that guide optimization processes, ensuring that generated responses align with human preferences and ethical considerations. Several studies have incorporated reward models into their RLHF frameworks, highlighting their significance in fine-tuning LLMs \citep{dong2024rlhf,bai2022training,ouyang2022training,dong2023raft,wang2024secrets}. Depending on the design of how to model the rewards, reward models can be categorized into both explicit and implicit frameworks. The former includes outcome-based or process-based ones \cite{ma2023let,zhang2025lessons,zhang2024generative,zhang2024rest,li2024process} with each assessing only the final output or intermediate steps respectively. In contrast, implicit reward modeling bypasses the need for explicit reward function engineering by leveraging preference optimization techniques. Notable examples include direct preference optimization (DPO) \cite{rafailov2024direct} and other variants like KTO \cite{ethayarajh2024kto}, SimPO \cite{meng2024simpo}, IPO \cite{azar2024general}, etc., all of which aim to refine LLM behavior without explicit reward function definition.
\section{Conclusion and Discussion}

Faced with the gap between existing answer selection methods and the upper bound from models' intrinsic abilities in the math problem-solving task, we propose \fans, a framework that introduces a formal language-based approach to perform better answer selection, offering a more rigorous and trustworthy alternative to heuristic methods like majority voting or best-of-N based on reward models. From the experiment results \ref{fig:main result}, we could see \fans indeed improve the accuracy for answer selection, achieving at most a relative improvement of $7.9\%$ on MATH500 Number Theory and $20\%$ on AMC23. This highlights the potential of utilizing formal language in enhancing answer selection for math problem-solving tasks. While it demonstrates notable improvements over baselines, there is still a gap between its current performance and the theoretical upper bound, i.e., pass @ N. Two primary directions for improvement stand out given the two-stage workflow of \fans.

\paragraph{Refining the NL-FL translation process is crucial to reducing false positives in theorem formulation.} Errors in translation, such as misrepresenting an equation-solving problem as a tautology or mishandling extremum problems by misinterpreting constraints, often lead to vacuous proofs and incorrect answer selection. Enhancing the translator to preserve problem constraints accurately and explicitly consider optimization conditions can significantly improve formalization quality. 
\paragraph{Enhancing the proving capabilities of the prover is necessary.} Lean4's current package ecosystem is biased toward certain fields like algebra and number theory, limiting its applicability in others like geometry and combinatorics. Expanding its library support would enable broader coverage of mathematical domains. Additionally, while the Goedel-Prover \cite{lin2024Goedelprover} achieves state-of-the-art results on formal language benchmarks \cite{zheng2021minif2f, ying2024lean}, it still struggles with nearly $40\%$ of proof problems. Strengthening its reasoning abilities through iterative fine-tuning and improved proof-search strategies could address this limitation.

Despite these challenges, the formal verification approach provides a trustworthy and interpretable alternative compared to conventional answer selection methods. By incorporating rigorously verifiable Lean4 deductions, \fans reduces dependence on extensive adaptation for different base models, ensuring consistent and scalable improvements in LLM-driven mathematical reasoning, shedding light on a more trustworthy way to better leverage the capability of existing models.
\section*{Limitations}

One major limitation of using formal language for answer selection is its high false positive rate and the inherent incompleteness of provers, which prohibits successfully verifying all factually correct statements. Errors in formalization, such as incorrect theorem representations or inadequate constraint handling, further lead to misclassification. Additionally, the limited domain coverage of Lean 4 constrains its applicability to only a subset of mathematical fields.

Future work could focus on improving the robustness of NL-FL translation, ensuring accurate problem formulation, and enhancing the theorem-proving capabilities of provers to handle more complex and diverse mathematical problems. Expanding the formal proof ecosystem beyond algebra and number theory will be essential for broader applicability as well. We believe addressing these challenges will bring formal verification closer to a more reliable, scalable solution for mathematical reasoning in LLMs.

For potential risks, since this work focuses on the math reasoning task of LLMs, with no intersection with ethics or other societal questions, we believe it currently contains no such risk. Besides, AI assistants were only used for spelling and grammar checking during writing.

\section*{Acknowledgments}
This research used both the DeltaAI advanced computing and data resource, which is supported by the National Science Foundation (award OAC 2320345) and the State of Illinois, and the Delta advanced computing and data resource which is supported by the National Science Foundation (award OAC 2005572) and the State of Illinois.. Delta and DeltaAI are joint efforts of the University of Illinois Urbana-Champaign and its National Center for Supercomputing Applications. This work is partially supported by NSF grant No. 2416897 and ONR grant No. N000142512318.

\newpage
\bibliography{content/ref}

\begin{thebibliography}{65}
\providecommand{\natexlab}[1]{#1}

\bibitem[{Azar et~al.(2024)Azar, Guo, Piot, Munos, Rowland, Valko, and Calandriello}]{azar2024general}
Mohammad~Gheshlaghi Azar, Zhaohan~Daniel Guo, Bilal Piot, Remi Munos, Mark Rowland, Michal Valko, and Daniele Calandriello. 2024.
\newblock A general theoretical paradigm to understand learning from human preferences.
\newblock In \emph{International Conference on Artificial Intelligence and Statistics}, pages 4447--4455. PMLR.

\bibitem[{Azerbayev et~al.(2023)Azerbayev, Schoelkopf, Paster, Santos, McAleer, Jiang, Deng, Biderman, and Welleck}]{azerbayev2023llemma}
Zhangir Azerbayev, Hailey Schoelkopf, Keiran Paster, Marco~Dos Santos, Stephen McAleer, Albert~Q Jiang, Jia Deng, Stella Biderman, and Sean Welleck. 2023.
\newblock Llemma: An open language model for mathematics.
\newblock \emph{arXiv preprint arXiv:2310.10631}.

\bibitem[{Bai et~al.(2022)Bai, Jones, Ndousse, Askell, Chen, DasSarma, Drain, Fort, Ganguli, Henighan et~al.}]{bai2022training}
Yuntao Bai, Andy Jones, Kamal Ndousse, Amanda Askell, Anna Chen, Nova DasSarma, Dawn Drain, Stanislav Fort, Deep Ganguli, Tom Henighan, and 1 others. 2022.
\newblock Training a helpful and harmless assistant with reinforcement learning from human feedback.
\newblock \emph{arXiv preprint arXiv:2204.05862}.

\bibitem[{Chen et~al.(2024)Chen, Wang, Cao, Liu, Gao, Cui, Zhu, Ye, Tian, Liu et~al.}]{chen2024expanding}
Zhe Chen, Weiyun Wang, Yue Cao, Yangzhou Liu, Zhangwei Gao, Erfei Cui, Jinguo Zhu, Shenglong Ye, Hao Tian, Zhaoyang Liu, and 1 others. 2024.
\newblock Expanding performance boundaries of open-source multimodal models with model, data, and test-time scaling.
\newblock \emph{arXiv preprint arXiv:2412.05271}.

\bibitem[{Cobbe et~al.(2021)Cobbe, Kosaraju, Bavarian, Chen, Jun, Kaiser, Plappert, Tworek, Hilton, Nakano et~al.}]{cobbe2021training}
Karl Cobbe, Vineet Kosaraju, Mohammad Bavarian, Mark Chen, Heewoo Jun, Lukasz Kaiser, Matthias Plappert, Jerry Tworek, Jacob Hilton, Reiichiro Nakano, and 1 others. 2021.
\newblock Training verifiers to solve math word problems.
\newblock \emph{arXiv preprint arXiv:2110.14168}.

\bibitem[{Coq(1996)}]{coq1996coq}
Projet Coq. 1996.
\newblock The coq proof assistant-reference manual.
\newblock \emph{INRIA Rocquencourt and ENS Lyon, version}, 5.

\bibitem[{De~Moura et~al.(2015)De~Moura, Kong, Avigad, Van~Doorn, and von Raumer}]{de2015lean}
Leonardo De~Moura, Soonho Kong, Jeremy Avigad, Floris Van~Doorn, and Jakob von Raumer. 2015.
\newblock The lean theorem prover (system description).
\newblock In \emph{Automated Deduction-CADE-25: 25th International Conference on Automated Deduction, Berlin, Germany, August 1-7, 2015, Proceedings 25}, pages 378--388. Springer.

\bibitem[{Dong et~al.(2023)Dong, Xiong, Goyal, Pan, Diao, Zhang, Shum, and Zhang}]{dong2023raft}
Hanze Dong, Wei Xiong, Deepanshu Goyal, Rui Pan, Shizhe Diao, Jipeng Zhang, Kashun Shum, and Tong Zhang. 2023.
\newblock Raft: Reward ranked finetuning for generative foundation model alignment.
\newblock \emph{arXiv preprint arXiv:2304.06767}.

\bibitem[{Dong et~al.(2024)Dong, Xiong, Pang, Wang, Zhao, Zhou, Jiang, Sahoo, Xiong, and Zhang}]{dong2024rlhf}
Hanze Dong, Wei Xiong, Bo~Pang, Haoxiang Wang, Han Zhao, Yingbo Zhou, Nan Jiang, Doyen Sahoo, Caiming Xiong, and Tong Zhang. 2024.
\newblock Rlhf workflow: From reward modeling to online rlhf.
\newblock \emph{arXiv preprint arXiv:2405.07863}.

\bibitem[{Ethayarajh et~al.(2024)Ethayarajh, Xu, Muennighoff, Jurafsky, and Kiela}]{ethayarajh2024kto}
Kawin Ethayarajh, Winnie Xu, Niklas Muennighoff, Dan Jurafsky, and Douwe Kiela. 2024.
\newblock Kto: Model alignment as prospect theoretic optimization.
\newblock \emph{arXiv preprint arXiv:2402.01306}.

\bibitem[{Guan et~al.(2025)Guan, Zhang, Liu, Shang, Sun, Zhu, Yang, and Yang}]{guan2025rstar}
Xinyu Guan, Li~Lyna Zhang, Yifei Liu, Ning Shang, Youran Sun, Yi~Zhu, Fan Yang, and Mao Yang. 2025.
\newblock rstar-math: Small llms can master math reasoning with self-evolved deep thinking.
\newblock \emph{arXiv preprint arXiv:2501.04519}.

\bibitem[{Harrison(2009)}]{harrison2009hol}
John Harrison. 2009.
\newblock Hol light: An overview.
\newblock In \emph{International Conference on Theorem Proving in Higher Order Logics}, pages 60--66. Springer.

\bibitem[{He et~al.(2024)He, Luo, Bai, Hu, Thai, Shen, Hu, Han, Huang, Zhang, Liu, Qi, Liu, and Sun}]{he2024olympiadbench}
Chaoqun He, Renjie Luo, Yuzhuo Bai, Shengding Hu, Zhen~Leng Thai, Junhao Shen, Jinyi Hu, Xu~Han, Yujie Huang, Yuxiang Zhang, Jie Liu, Lei Qi, Zhiyuan Liu, and Maosong Sun. 2024.
\newblock \href {https://arxiv.org/abs/2402.14008} {Olympiadbench: A challenging benchmark for promoting agi with olympiad-level bilingual multimodal scientific problems}.
\newblock \emph{Preprint}, arXiv:2402.14008.

\bibitem[{Hendrycks et~al.(2021)Hendrycks, Burns, Kadavath, Arora, Basart, Tang, Song, and Steinhardt}]{hendrycks2021measuring}
Dan Hendrycks, Collin Burns, Saurav Kadavath, Akul Arora, Steven Basart, Eric Tang, Dawn Song, and Jacob Steinhardt. 2021.
\newblock Measuring mathematical problem solving with the math dataset.
\newblock \emph{arXiv preprint arXiv:2103.03874}.

\bibitem[{Jiang et~al.(2023)Jiang, Sablayrolles, Mensch, Bamford, Chaplot, Casas, Bressand, Lengyel, Lample, Saulnier et~al.}]{jiang2023mistral}
Albert~Q Jiang, Alexandre Sablayrolles, Arthur Mensch, Chris Bamford, Devendra~Singh Chaplot, Diego de~las Casas, Florian Bressand, Gianna Lengyel, Guillaume Lample, Lucile Saulnier, and 1 others. 2023.
\newblock Mistral 7b.
\newblock \emph{arXiv preprint arXiv:2310.06825}.

\bibitem[{Jiang et~al.(2022)Jiang, Welleck, Zhou, Li, Liu, Jamnik, Lacroix, Wu, and Lample}]{jiang2022draft}
Albert~Q Jiang, Sean Welleck, Jin~Peng Zhou, Wenda Li, Jiacheng Liu, Mateja Jamnik, Timoth{\'e}e Lacroix, Yuhuai Wu, and Guillaume Lample. 2022.
\newblock Draft, sketch, and prove: Guiding formal theorem provers with informal proofs.
\newblock \emph{arXiv preprint arXiv:2210.12283}.

\bibitem[{Jiang et~al.(2024)Jiang, Chen, Min, Chen, Cheng, Wang, Tang, Sun, Deng, Zhao et~al.}]{jiang2024technical}
Jinhao Jiang, Zhipeng Chen, Yingqian Min, Jie Chen, Xiaoxue Cheng, Jiapeng Wang, Yiru Tang, Haoxiang Sun, Jia Deng, Wayne~Xin Zhao, and 1 others. 2024.
\newblock Technical report: Enhancing llm reasoning with reward-guided tree search.
\newblock \emph{arXiv preprint arXiv:2411.11694}.

\bibitem[{{Kimi Team}(2025)}]{MoonshotAI}
{Kimi Team}. 2025.
\newblock Kimi k1.5: Scaling reinforcement learning with llms.

\bibitem[{Kwon et~al.(2023)Kwon, Li, Zhuang, Sheng, Zheng, Yu, Gonzalez, Zhang, and Stoica}]{kwon2023efficient}
Woosuk Kwon, Zhuohan Li, Siyuan Zhuang, Ying Sheng, Lianmin Zheng, Cody~Hao Yu, Joseph~E. Gonzalez, Hao Zhang, and Ion Stoica. 2023.
\newblock Efficient memory management for large language model serving with pagedattention.
\newblock In \emph{Proceedings of the ACM SIGOPS 29th Symposium on Operating Systems Principles}.

\bibitem[{Lewkowycz et~al.(2022)Lewkowycz, Andreassen, Dohan, Dyer, Michalewski, Ramasesh, Slone, Anil, Schlag, Gutman-Solo et~al.}]{lewkowycz2022solving}
Aitor Lewkowycz, Anders Andreassen, David Dohan, Ethan Dyer, Henryk Michalewski, Vinay Ramasesh, Ambrose Slone, Cem Anil, Imanol Schlag, Theo Gutman-Solo, and 1 others. 2022.
\newblock Solving quantitative reasoning problems with language models.
\newblock \emph{Advances in Neural Information Processing Systems}, 35:3843--3857.

\bibitem[{Li and Li(2024)}]{li2024process}
Wendi Li and Yixuan Li. 2024.
\newblock Process reward model with q-value rankings.
\newblock \emph{arXiv preprint arXiv:2410.11287}.

\bibitem[{Lightman et~al.(2023)Lightman, Kosaraju, Burda, Edwards, Baker, Lee, Leike, Schulman, Sutskever, and Cobbe}]{lightman2023lets}
Hunter Lightman, Vineet Kosaraju, Yura Burda, Harri Edwards, Bowen Baker, Teddy Lee, Jan Leike, John Schulman, Ilya Sutskever, and Karl Cobbe. 2023.
\newblock Let's verify step by step.
\newblock \emph{arXiv preprint arXiv:2305.20050}.

\bibitem[{Lin et~al.(2024)Lin, Sun, Yang, and Welleck}]{lin2024lean}
Haohan Lin, Zhiqing Sun, Yiming Yang, and Sean Welleck. 2024.
\newblock Lean-star: Learning to interleave thinking and proving.
\newblock \emph{arXiv preprint arXiv:2407.10040}.

\bibitem[{Lin et~al.(2025)Lin, Tang, Lyu, Wu, Lin, Yang, Li, Xia, Chen, Arora, and Jin}]{lin2024Goedelprover}
Yong Lin, Shange Tang, Bohan Lyu, Jiayun Wu, Hongzhou Lin, Kaiyu Yang, Jia Li, Mengzhou Xia, Danqi Chen, Sanjeev Arora, and Chi Jin. 2025.
\newblock \href {https://arxiv.org/abs/2502.07640} {Goedel-prover: A frontier model for open-source automated theorem proving}.
\newblock \emph{Preprint}, arXiv:2502.07640.

\bibitem[{Ma et~al.(2023)Ma, Zhou, Liu, Yuan, Liu, You, and Yang}]{ma2023let}
Qianli Ma, Haotian Zhou, Tingkai Liu, Jianbo Yuan, Pengfei Liu, Yang You, and Hongxia Yang. 2023.
\newblock Let's reward step by step: Step-level reward model as the navigators for reasoning.
\newblock \emph{arXiv preprint arXiv:2310.10080}.

\bibitem[{Megill and Wheeler(2019)}]{megill2019metamath}
Norman Megill and David~A Wheeler. 2019.
\newblock \emph{Metamath: a computer language for mathematical proofs}.
\newblock Lulu. com.

\bibitem[{Meng et~al.(2024)Meng, Xia, and Chen}]{meng2024simpo}
Yu~Meng, Mengzhou Xia, and Danqi Chen. 2024.
\newblock Simpo: Simple preference optimization with a reference-free reward.
\newblock \emph{arXiv preprint arXiv:2405.14734}.

\bibitem[{Moura and Ullrich(2021)}]{moura2021lean}
Leonardo~de Moura and Sebastian Ullrich. 2021.
\newblock The lean 4 theorem prover and programming language.
\newblock In \emph{Automated Deduction--CADE 28: 28th International Conference on Automated Deduction, Virtual Event, July 12--15, 2021, Proceedings 28}, pages 625--635. Springer.

\bibitem[{Muennighoff et~al.(2025)Muennighoff, Yang, Shi, Li, Fei-Fei, Hajishirzi, Zettlemoyer, Liang, Cand{\`e}s, and Hashimoto}]{muennighoff2025s1}
Niklas Muennighoff, Zitong Yang, Weijia Shi, Xiang~Lisa Li, Li~Fei-Fei, Hannaneh Hajishirzi, Luke Zettlemoyer, Percy Liang, Emmanuel Cand{\`e}s, and Tatsunori Hashimoto. 2025.
\newblock s1: Simple test-time scaling.
\newblock \emph{arXiv preprint arXiv:2501.19393}.

\bibitem[{Open-Source-O1(2024)}]{open_o1}
Open-Source-O1. 2024.
\newblock \href {https://github.com/Open-Source-O1/Open-O1} {Open-o1}.
\newblock Accessed: 2024-12-28.

\bibitem[{{OpenAI}(2024)}]{openai2024reasoning}
{OpenAI}. 2024.
\newblock Learning to reason with llms.
\newblock \url{https://openai.com/index/learning-to-reason-with-llms/}.
\newblock Accessed: 2024-11-24.

\bibitem[{Ouyang et~al.(2022)Ouyang, Wu, Jiang, Almeida, Wainwright, Mishkin, Zhang, Agarwal, Slama, Ray et~al.}]{ouyang2022training}
Long Ouyang, Jeffrey Wu, Xu~Jiang, Diogo Almeida, Carroll Wainwright, Pamela Mishkin, Chong Zhang, Sandhini Agarwal, Katarina Slama, Alex Ray, and 1 others. 2022.
\newblock Training language models to follow instructions with human feedback.
\newblock \emph{Advances in neural information processing systems}, 35:27730--27744.

\bibitem[{Paulson(1994)}]{paulson1994isabelle}
Lawrence~C Paulson. 1994.
\newblock \emph{Isabelle: A generic theorem prover}.
\newblock Springer.

\bibitem[{{Qwen Team}(2025)}]{qwq32b}
{Qwen Team}. 2025.
\newblock \href {https://qwenlm.github.io/blog/qwq-32b/} {Qwq-32b: Embracing the power of reinforcement learning}.

\bibitem[{Rafailov et~al.(2024)Rafailov, Sharma, Mitchell, Manning, Ermon, and Finn}]{rafailov2024direct}
Rafael Rafailov, Archit Sharma, Eric Mitchell, Christopher~D Manning, Stefano Ermon, and Chelsea Finn. 2024.
\newblock Direct preference optimization: Your language model is secretly a reward model.
\newblock \emph{Advances in Neural Information Processing Systems}, 36.

\bibitem[{Ren et~al.(2025)Ren, Shao, Song, Xin, Wang, Zhao, Zhang, Fu, Zhu, Yang et~al.}]{ren2025deepseek}
ZZ~Ren, Zhihong Shao, Junxiao Song, Huajian Xin, Haocheng Wang, Wanjia Zhao, Liyue Zhang, Zhe Fu, Qihao Zhu, Dejian Yang, and 1 others. 2025.
\newblock Deepseek-prover-v2: Advancing formal mathematical reasoning via reinforcement learning for subgoal decomposition.
\newblock \emph{arXiv preprint arXiv:2504.21801}.

\bibitem[{Santos et~al.(2025)Santos, Wang, de~Saxcé, Wang, Baksys, Unsal, Liu, Liu, and Li}]{santos2025kiminaleanservertechnical}
Marco~Dos Santos, Haiming Wang, Hugues de~Saxcé, Ran Wang, Mantas Baksys, Mert Unsal, Junqi Liu, Zhengying Liu, and Jia Li. 2025.
\newblock \href {https://arxiv.org/abs/2504.21230} {Kimina lean server: Technical report}.
\newblock \emph{Preprint}, arXiv:2504.21230.

\bibitem[{Saparov et~al.(2023)Saparov, Pang, Padmakumar, Joshi, Kazemi, Kim, and He}]{saparov2023testing}
Abulhair Saparov, Richard~Yuanzhe Pang, Vishakh Padmakumar, Nitish Joshi, Mehran Kazemi, Najoung Kim, and He~He. 2023.
\newblock Testing the general deductive reasoning capacity of large language models using ood examples.
\newblock \emph{Advances in Neural Information Processing Systems}, 36:3083--3105.

\bibitem[{Shao et~al.(2024)Shao, Wang, Zhu, Xu, Song, Bi, Zhang, Zhang, Li, Wu et~al.}]{shao2024deepseekmath}
Zhihong Shao, Peiyi Wang, Qihao Zhu, Runxin Xu, Junxiao Song, Xiao Bi, Haowei Zhang, Mingchuan Zhang, YK~Li, Y~Wu, and 1 others. 2024.
\newblock Deepseekmath: Pushing the limits of mathematical reasoning in open language models.
\newblock \emph{arXiv preprint arXiv:2402.03300}.

\bibitem[{Snell et~al.(2024)Snell, Lee, Xu, and Kumar}]{snell2024scaling}
Charlie Snell, Jaehoon Lee, Kelvin Xu, and Aviral Kumar. 2024.
\newblock Scaling llm test-time compute optimally can be more effective than scaling model parameters.
\newblock \emph{arXiv preprint arXiv:2408.03314}.

\bibitem[{Wang et~al.(2024{\natexlab{a}})Wang, Zheng, Chen, Liu, Dou, Huang, Shen, Jin, Zhou, Shi et~al.}]{wang2024secrets}
Binghai Wang, Rui Zheng, Lu~Chen, Yan Liu, Shihan Dou, Caishuang Huang, Wei Shen, Senjie Jin, Enyu Zhou, Chenyu Shi, and 1 others. 2024{\natexlab{a}}.
\newblock Secrets of rlhf in large language models part ii: Reward modeling.
\newblock \emph{arXiv preprint arXiv:2401.06080}.

\bibitem[{Wang et~al.(2025{\natexlab{a}})Wang, Unsal, Lin, Baksys, Liu, Santos, Sung, Vinyes, Ying, Zhu et~al.}]{wang2025kimina}
Haiming Wang, Mert Unsal, Xiaohan Lin, Mantas Baksys, Junqi Liu, Marco~Dos Santos, Flood Sung, Marina Vinyes, Zhenzhe Ying, Zekai Zhu, and 1 others. 2025{\natexlab{a}}.
\newblock Kimina-prover preview: Towards large formal reasoning models with reinforcement learning.
\newblock \emph{arXiv preprint arXiv:2504.11354}.

\bibitem[{Wang et~al.(2025{\natexlab{b}})Wang, Pan, Li, Zhang, Jia, Diao, Pi, Hu, and Zhang}]{wang2025malot}
Ruida Wang, Rui Pan, Yuxin Li, Jipeng Zhang, Yizhen Jia, Shizhe Diao, Renjie Pi, Junjie Hu, and Tong Zhang. 2025{\natexlab{b}}.
\newblock Ma-lot: Multi-agent lean-based long chain-of-thought reasoning enhances formal theorem proving.
\newblock \emph{arXiv preprint arXiv:2503.03205}.

\bibitem[{Wang et~al.(2024{\natexlab{b}})Wang, Zhang, Jia, Pan, Diao, Pi, and Zhang}]{wang2024theoremllama}
Ruida Wang, Jipeng Zhang, Yizhen Jia, Rui Pan, Shizhe Diao, Renjie Pi, and Tong Zhang. 2024{\natexlab{b}}.
\newblock Theoremllama: Transforming general-purpose llms into lean4 experts.
\newblock \emph{arXiv preprint arXiv:2407.03203}.

\bibitem[{Wang et~al.(2023)Wang, Zhou, and Sachan}]{wang2023let}
Ruida Wang, Wangchunshu Zhou, and Mrinmaya Sachan. 2023.
\newblock Let's synthesize step by step: Iterative dataset synthesis with large language models by extrapolating errors from small models.
\newblock \emph{arXiv preprint arXiv:2310.13671}.

\bibitem[{Wei et~al.(2022)Wei, Wang, Schuurmans, Bosma, Xia, Chi, Le, Zhou et~al.}]{wei2022chain}
Jason Wei, Xuezhi Wang, Dale Schuurmans, Maarten Bosma, Fei Xia, Ed~Chi, Quoc~V Le, Denny Zhou, and 1 others. 2022.
\newblock Chain-of-thought prompting elicits reasoning in large language models.
\newblock \emph{Advances in neural information processing systems}, 35:24824--24837.

\bibitem[{Wolf et~al.(2020)Wolf, Debut, Sanh, Chaumond, Delangue, Moi, Cistac, Rault, Louf, Funtowicz, Davison, Shleifer, von Platen, Ma, Jernite, Plu, Xu, Scao, Gugger, Drame, Lhoest, and Rush}]{wolf-etal-2020-transformers}
Thomas Wolf, Lysandre Debut, Victor Sanh, Julien Chaumond, Clement Delangue, Anthony Moi, Pierric Cistac, Tim Rault, Rémi Louf, Morgan Funtowicz, Joe Davison, Sam Shleifer, Patrick von Platen, Clara Ma, Yacine Jernite, Julien Plu, Canwen Xu, Teven~Le Scao, Sylvain Gugger, and 3 others. 2020.
\newblock \href {https://www.aclweb.org/anthology/2020.emnlp-demos.6} {Transformers: State-of-the-art natural language processing}.
\newblock In \emph{Proceedings of the 2020 Conference on Empirical Methods in Natural Language Processing: System Demonstrations}, pages 38--45, Online. Association for Computational Linguistics.

\bibitem[{Wu et~al.(2024{\natexlab{a}})Wu, Sun, Li, Welleck, and Yang}]{wu2024inference}
Yangzhen Wu, Zhiqing Sun, Shanda Li, Sean Welleck, and Yiming Yang. 2024{\natexlab{a}}.
\newblock Inference scaling laws: An empirical analysis of compute-optimal inference for problem-solving with language models.
\newblock \emph{arXiv preprint arXiv:2408.00724}.

\bibitem[{Wu et~al.(2024{\natexlab{b}})Wu, Huang, Zhou, Ying, Wang, Lin, and Chen}]{wu2024internlm2}
Zijian Wu, Suozhi Huang, Zhejian Zhou, Huaiyuan Ying, Jiayu Wang, Dahua Lin, and Kai Chen. 2024{\natexlab{b}}.
\newblock Internlm2. 5-stepprover: Advancing automated theorem proving via expert iteration on large-scale lean problems.
\newblock \emph{arXiv preprint arXiv:2410.15700}.

\bibitem[{Xin et~al.(2024{\natexlab{a}})Xin, Guo, Shao, Ren, Zhu, Liu, Ruan, Li, and Liang}]{xin2024deepseek1}
Huajian Xin, Daya Guo, Zhihong Shao, Zhizhou Ren, Qihao Zhu, Bo~Liu, Chong Ruan, Wenda Li, and Xiaodan Liang. 2024{\natexlab{a}}.
\newblock Deepseek-prover: Advancing theorem proving in llms through large-scale synthetic data.
\newblock \emph{arXiv preprint arXiv:2405.14333}.

\bibitem[{Xin et~al.(2024{\natexlab{b}})Xin, Ren, Song, Shao, Zhao, Wang, Liu, Zhang, Lu, Du et~al.}]{xin2024deepseek}
Huajian Xin, ZZ~Ren, Junxiao Song, Zhihong Shao, Wanjia Zhao, Haocheng Wang, Bo~Liu, Liyue Zhang, Xuan Lu, Qiushi Du, and 1 others. 2024{\natexlab{b}}.
\newblock Deepseek-prover-v1. 5: Harnessing proof assistant feedback for reinforcement learning and monte-carlo tree search.
\newblock \emph{arXiv preprint arXiv:2408.08152}.

\bibitem[{Xiong et~al.(2025)Xiong, Yao, Xu, Pang, Wang, Sahoo, Li, Jiang, Zhang, Xiong et~al.}]{xiong2025minimalist}
Wei Xiong, Jiarui Yao, Yuhui Xu, Bo~Pang, Lei Wang, Doyen Sahoo, Junnan Li, Nan Jiang, Tong Zhang, Caiming Xiong, and 1 others. 2025.
\newblock A minimalist approach to llm reasoning: from rejection sampling to reinforce.
\newblock \emph{arXiv preprint arXiv:2504.11343}.

\bibitem[{Yang et~al.(2024{\natexlab{a}})Yang, Yang, Hui, Zheng, Yu, Zhou, Li, Li, Liu, Huang et~al.}]{yang2024qwen2}
An~Yang, Baosong Yang, Binyuan Hui, Bo~Zheng, Bowen Yu, Chang Zhou, Chengpeng Li, Chengyuan Li, Dayiheng Liu, Fei Huang, and 1 others. 2024{\natexlab{a}}.
\newblock Qwen2 technical report.
\newblock \emph{arXiv preprint arXiv:2407.10671}.

\bibitem[{Yang et~al.(2024{\natexlab{b}})Yang, Zhang, Hui, Gao, Yu, Li, Liu, Tu, Zhou, Lin, Lu, Xue, Lin, Liu, Ren, and Zhang}]{yang2024qwen25mathtechnicalreportmathematical}
An~Yang, Beichen Zhang, Binyuan Hui, Bofei Gao, Bowen Yu, Chengpeng Li, Dayiheng Liu, Jianhong Tu, Jingren Zhou, Junyang Lin, Keming Lu, Mingfeng Xue, Runji Lin, Tianyu Liu, Xingzhang Ren, and Zhenru Zhang. 2024{\natexlab{b}}.
\newblock Qwen2.5-math technical report: Toward mathematical expert model via self-improvement.
\newblock \emph{arXiv preprint arXiv:2409.12122}.

\bibitem[{Yang et~al.(2024{\natexlab{c}})Yang, Swope, Gu, Chalamala, Song, Yu, Godil, Prenger, and Anandkumar}]{yang2024leandojo}
Kaiyu Yang, Aidan Swope, Alex Gu, Rahul Chalamala, Peiyang Song, Shixing Yu, Saad Godil, Ryan~J Prenger, and Animashree Anandkumar. 2024{\natexlab{c}}.
\newblock Leandojo: Theorem proving with retrieval-augmented language models.
\newblock \emph{Advances in Neural Information Processing Systems}, 36.

\bibitem[{Yao et~al.(2025)Yao, Hao, Zhang, Dong, Xiong, Jiang, and Zhang}]{yao2025optimizing}
Jiarui Yao, Yifan Hao, Hanning Zhang, Hanze Dong, Wei Xiong, Nan Jiang, and Tong Zhang. 2025.
\newblock Optimizing chain-of-thought reasoners via gradient variance minimization in rejection sampling and rl.
\newblock \emph{arXiv preprint arXiv:2505.02391}.

\bibitem[{Yao et~al.(2023)Yao, Yu, Zhao, Shafran, Griffiths, Cao, and Narasimhan}]{yao2023tree}
Shunyu Yao, Dian Yu, Jeffrey Zhao, Izhak Shafran, Thomas~L Griffiths, Yuan Cao, and Karthik Narasimhan. 2023.
\newblock Tree of thoughts: Deliberate problem solving with large language models, 2023.
\newblock \emph{URL https://arxiv. org/pdf/2305.10601. pdf}.

\bibitem[{Ying et~al.(2024)Ying, Wu, Geng, Wang, Lin, and Chen}]{ying2024lean}
Huaiyuan Ying, Zijian Wu, Yihan Geng, Jiayu Wang, Dahua Lin, and Kai Chen. 2024.
\newblock Lean workbook: A large-scale lean problem set formalized from natural language math problems.
\newblock \emph{arXiv preprint arXiv:2406.03847}.

\bibitem[{Yu et~al.(2023)Yu, Jiang, Shi, Yu, Liu, Zhang, Kwok, Li, Weller, and Liu}]{yu2023metamath}
Longhui Yu, Weisen Jiang, Han Shi, Jincheng Yu, Zhengying Liu, Yu~Zhang, James~T Kwok, Zhenguo Li, Adrian Weller, and Weiyang Liu. 2023.
\newblock Metamath: Bootstrap your own mathematical questions for large language models.
\newblock \emph{arXiv preprint arXiv:2309.12284}.

\bibitem[{Zhang et~al.(2024{\natexlab{a}})Zhang, Zhoubian, Hu, Yue, Dong, and Tang}]{zhang2024rest}
Dan Zhang, Sining Zhoubian, Ziniu Hu, Yisong Yue, Yuxiao Dong, and Jie Tang. 2024{\natexlab{a}}.
\newblock Rest-mcts*: Llm self-training via process reward guided tree search.
\newblock \emph{arXiv preprint arXiv:2406.03816}.

\bibitem[{Zhang et~al.(2025{\natexlab{a}})Zhang, Yao, Ye, Xiong, and Zhang}]{zhang2025online}
Hanning Zhang, Jiarui Yao, Chenlu Ye, Wei Xiong, and Tong Zhang. 2025{\natexlab{a}}.
\newblock Online-dpo-r1: Unlocking effective reasoning without the ppo overhead.

\bibitem[{Zhang et~al.(2024{\natexlab{b}})Zhang, Hosseini, Bansal, Kazemi, Kumar, and Agarwal}]{zhang2024generative}
Lunjun Zhang, Arian Hosseini, Hritik Bansal, Mehran Kazemi, Aviral Kumar, and Rishabh Agarwal. 2024{\natexlab{b}}.
\newblock Generative verifiers: Reward modeling as next-token prediction.
\newblock \emph{arXiv preprint arXiv:2408.15240}.

\bibitem[{Zhang et~al.(2025{\natexlab{b}})Zhang, Zheng, Wu, Zhang, Lin, Yu, Liu, Zhou, and Lin}]{zhang2025lessons}
Zhenru Zhang, Chujie Zheng, Yangzhen Wu, Beichen Zhang, Runji Lin, Bowen Yu, Dayiheng Liu, Jingren Zhou, and Junyang Lin. 2025{\natexlab{b}}.
\newblock The lessons of developing process reward models in mathematical reasoning.
\newblock \emph{arXiv preprint arXiv:2501.07301}.

\bibitem[{Zheng et~al.(2021)Zheng, Han, and Polu}]{zheng2021minif2f}
Kunhao Zheng, Jesse~Michael Han, and Stanislas Polu. 2021.
\newblock Minif2f: a cross-system benchmark for formal olympiad-level mathematics.
\newblock \emph{arXiv preprint arXiv:2109.00110}.

\bibitem[{Zhou et~al.(2024)Zhou, Staats, Li, Szegedy, Weinberger, and Wu}]{zhou2024don}
Jin~Peng Zhou, Charles Staats, Wenda Li, Christian Szegedy, Kilian~Q Weinberger, and Yuhuai Wu. 2024.
\newblock Don't trust: Verify--grounding llm quantitative reasoning with autoformalization.
\newblock \emph{arXiv preprint arXiv:2403.18120}.

\end{thebibliography}

\appendix
\label{sec:appendix}


\section{Translator Training Example}\label{appendix:translatorExample}

This section gives examples of the training and inference method for using the transfer learning method to train the Long CoT capability of translator based on data without special Long CoT annotation.

The prompt for training is as follows:

\begin{Verbatim}[breaklines=True]
[
  {   
    "role": "system", 
    "content":"You are a helpful mathematical assistant specialized in formal language Lean4. Please translate the following natural language statement into Lean4 theorem statement. You are directly asked to provide the answer to the problem **WITHOUT** using the Long Chain of Thought."
  }, 
  {
    "role": "user", 
    "content": """@ Natural language theorem statement: Let $a,b,c$ be positive real numbers .Prove that $ \\frac{b+c}{\\sqrt{a^2 + 8bc}} + \\frac{c+a}{\\sqrt{b^2 + 8ca}} + \\frac{a+b}{\\sqrt{c^2 + 8ab}} \\geq 2. $

    @ Lean4 theorem statement:"""
  }, 
  {
    "role": "assistant", 
    "content": """<Thought> The user ask not to solve with long CoT, so I will directly write the answer.</Thought>

    ```lean4
    theorem lean_workbook_0 (a b c : R) (ha : 0 < a) (hb : 0 < b) (hc : 0 < c) : (b + c) / Real.sqrt (a ^ 2 + 8 * b * c) + (c + a) / Real.sqrt (b ^ 2 + 8 * c * a) + (a + b) / Real.sqrt (c ^ 2 + 8 * a * b) >= 2  :=  by
    ```"""
  }
]  
\end{Verbatim}

We use system prompt and user input as the training input and the assistant part as the expected output for our model training. We can see that in the system prompt, we indicate the model not to use Long CoT and in training, we just put a dummy placeholder for the Long CoT. 

The inference output is as follows:

\begin{Verbatim}[breaklines=True]
===Model Input===
<|begin_of_sentence|>Follow these instructions carefully:
1. Provide a logically correct and rigorous translate of natural language theorem statement to Lean4 statement.
2. In the <Thought> section, include your detailed step-by-step reasoning.
3. In the <Output> section, provide only the final Lean4 proof or final result.### Instruction:You will receive several Lean4 problems. For each:
- **Use** a step-by-step solution internally in <Thought>.
- **Do not** reveal your chain of thought outside the <Thought> block.
- **Ensure** the final Lean4 code or final result is placed **only** in <Output>.
@ Natural language theorem statement:
mathd_numbertheory_3
What is the units digit of the sum of the squares of the first nine positive integers? Show that it is 5.

@ Lean4 theorem statement:
### Response:
<Thought>

Alright, I should do the following:

  1. Analysis the problem and understand the requirements.

  2. Use natural language to analysis the potentional Lean4 condition and requirements in the natural language theorem statement.

  3. Write the translated Lean4 statement code.

===Model Output===
he `mathd_numbertheory_3` has the following conditions:

- The problem asks for the units digit of the sum of the squares of the first nine positive integers.
- The expected answer is 5.

So, the theorem should state that the units digit of the sum of the squares of the first nine positive integers is 5.
</Thought>

<Output>
```lean4
theorem mathd_numbertheory_3 : (Sum i in Finset.range 10, i ^ 2) % 10 = 5 :=
```&
</Output>
\end{Verbatim}

We can see that by using the system prompt to clearly instruct the model to use Long CoT in inference, the model apply its internal bearing Long CoT capability, together with the Lean4 knowledge and statement translation ability learned in transfer learning. The model can perform accurate translation from NL to FL using Long CoT inference.

\section{Experiments Details}
\subsection{Implementation Settings}
For the answer generation, we use vLLM \cite{kwon2023efficient} as the inference framework, with the temperature set to 0.6 to avoid too much repetition, and set the max tokens allowed to be generated as 4096.
During the first stage of translation, we set the temperature of generation to 0, together with the number of max new tokens set to 2048, and use Transformers \cite{wolf-etal-2020-transformers} as the inference framework directly. For the prover, we use vLLM for inference as well and set the temperature to 1.0, with the max number of new tokens set to 2048.

For all experiment results, we evaluate our method \fans on four different base models, Mistral-7B \cite{jiang2023mistral}, DeepSeek-Math-7B-Instruct \cite{shao2024deepseekmath}, Qwen-2.5-Math-1.5B-Instruct \cite{yang2024qwen25mathtechnicalreportmathematical}, Qwen2-Math-7B-Instruct \cite{yang2024qwen2}, and the results could be found in table \ref{tab:full experiment}. We did not include Qwen-2.5-Math-7B-Instruct as one of our base models because during inference we found that it could easily generate nonsense outputs not related to the math problems until the inference budget is reached. This phenomenon persists no matter what inference framework we use. So as an alternative, we chose the 7B model from a version before. Overall, the performance of the latter two models is better than the former two. And \fans could achieve uniform performance gain among all different base models, with a larger margin on weaker base models. 

The performance gain on stronger base models is smaller due to the fact that harder problems are more challenging to be translated into appropriate natural language statements with consistent meaning and math formulation, and they are hard to be proved automatically by the prover in stage two. In an ideal situation, if the success rates for NL-FL translation and prover are $p$ and $q$ respectively, and the accuracies for majority vote and pass @ n are $r_1$ and $r_2$ respectively, then the theoretical performance gain of \fans should be roughly $(r_2-r_1)\cdot pq$. This leaves a huge space for future work to improve the current pipeline from both the point of views of translator and prover. With better translator achieving lower false positive rates, and / or better prover which could systematically prove harder formal language statements, \fans will be more effective and efficient benefiting from both aspects.

\begin{table*}[]
    \centering
    \begin{adjustbox}{max width=\textwidth}\begin{tabular}{llccccccc}
    \toprule
        Base Model & Method & Precalculus & Prealgebra & InterAlgebra & Algebra & Number Theory & MATH500 & AMC23\\
        \midrule
        \multirow{4}{*}{Mistral-7B} & Pass @ 1 & 0.1071 & 0.4954 & 0.1289 & 0.3579 & 0.2681 & 0.2785 & 0.0781\\
        & Pass @ 8 & 0.3214 & 0.8049 & 0.3402 & 0.6532 & 0.4677 & 0.5180 & 0.3250\\
        & Majority Vote @ 8 & \textbf{0.1786} & 0.5732 & 0.1546 & 0.4274 & 0.3387 & \textbf{0.3880} & 0.1250 \\
        & FANS @ 8 & \textbf{0.1786} & \textbf{0.6341} & \textbf{0.1753} & \textbf{0.4597} & \textbf{0.3548} & 0.3640 & \textbf{0.1500} \\
        \midrule 
        \multirow{6}{*}{Deepseek-Math-7B-Instruct} & Pass @ 1 &  0.1696 & 0.6387 & 0.2023 & 0.5998 & 0.3609 & 0.4098 & 0.1687\\
        & Pass @ 8  & 0.4107 & 0.8780 & 0.4742 & 0.8548 & 0.7258 & 0.6860 & 0.4500\\
        & Majority Vote @ 8 & 0.1786 & 0.7195 & 0.3402 & 0.7177 & 0.5484 & 0.5200 & 0.2250 \\
        & FANS @ 8 & 0.1786 & 0.7195 & {0.3608} & {0.7339} & 0.5484 & {0.5240} & 0.2250\\
        & ORM @ 8 & \textbf{0.3750} & 0.8049 & 0.4021 & \textbf{0.8226} & 0.6129 & 0.6260 & 0.3000\\
        & ORM + FANS @ 8 &\textbf{0.3750} & \textbf{0.8171} & \textbf{0.4227} & \textbf{0.8226} & \textbf{0.6613} & \textbf{0.6380} & \textbf{0.3250} \\
        \midrule 
        \multirow{6}{*}{Qwen-2.5-Math-1.5B-Instruct} & Pass @ 1 & 0.5558 & 0.7774 & 0.5296 & 0.8871 & 0.8165 & 0.7055 & 0.4938 \\
        &Pass @ 8 & 0.7500 & 0.8902 & 0.7526 & 0.9839 & 0.9839 & 0.8720 & 0.8000 \\
        &Majority Vote @ 8 & 0.6250 & 0.8415 & 0.5979 & 0.9435 & 0.8871 & 0.7740 & 0.6000 \\
        &FANS @ 8 & 0.6250 & 0.8659 & 0.5979 & 0.9435 & 0.8871 & 0.7760 & 0.6250 \\
        &ORM @ 8 & \textbf{0.6607} & \textbf{0.8780} & 0.6392 & 0.9677 & 0.9194 & 0.8080 & \textbf{0.7000} \\
        &ORM + FANS @ 8 & \textbf{0.6607} & \textbf{0.8780} & \textbf{0.6495} & \textbf{0.9839} & \textbf{0.9355} & \textbf{0.8180} & \textbf{0.7000} \\
    \midrule
    \multirow{6}{*}{Qwen-2-Math-7B-Instruct} & Pass @ 1 & 0.5446 & 0.8110 & 0.5206 & 0.8891 & 0.8145 & 0.7182 & 0.4781 \\
        & Pass @ 8 & 0.7321 & 0.9302 & 0.7320 & 0.9758 & 0.9677 & 0.8700 & 0.8500 \\
        & Majority Vote @ 8 & 0.6071 & 0.8780 & 0.5773 & 0.9355 & 0.9194 & 0.7840 & 0.5500 \\
        & FANS @ 8 & 0.5893 & 0.8659 & 0.5773 & 0.9355 & 0.9194 & 0.7760 & 0.5750 \\
        & ORM @ 8 & \textbf{0.6964} & \textbf{0.8902} & 0.6289 & \textbf{0.9597} & 0.9032 & 0.8180 & \textbf{0.7500} \\
        & ORM + FANS @ 8 & \textbf{0.6964} & \textbf{0.8902} & \textbf{0.6392} & \textbf{0.9597} & \textbf{0.9194} & \textbf{0.8220} & \textbf{0.7500} \\
    \bottomrule
    \end{tabular}
    \end{adjustbox}
    \caption{Full experiments results.}
    \label{tab:full experiment}
\end{table*}

\begin{table*}[]
    \centering
    \begin{adjustbox}{max width=\textwidth}
    \begin{tabular}{lcccccccc}
        \toprule
        Method & Precalculus & Prealgebra & InterAlgebra & Algebra & {Number Theory} & Geometry & Counting \& Probability & MATH500 \\
        \midrule
        \fans @ 8 & \textbf{0.2143} & \textbf{0.5976} & 0.1443 & 0.4839 & \textbf{0.3710} & 0.2195 & \textbf{0.2632} & \textbf{0.354} \\
        \fans @ 8 aligned NL & 0.1786 & 0.5244 & \textbf{0.1753} & \textbf{0.5081} & 0.3548 & \textbf{0.3171} & 0.2368 & \textbf{0.354} \\
        \bottomrule
    \end{tabular}
    \end{adjustbox}
    \caption{Comparision between FANS with aligned natural language statement and original ones.}
    \label{tab:compare trans nl}
\end{table*}

\subsection{Effects of Natural Language Alignment}
Without natural language statements alignment, we directly append the generated answers to the end of the questions and utilize the concatenated ones as instructions for the formal language translator. In this scenario, the prompt used for formal language translation is table \ref{translation prompt no align}.
\begin{figure}[]
\centering
\begin{tcolorbox}[title={Formal Language Translation Prompt without Natural Language Alignment}]
    "\{question\} Show that it is: \{answer\}."
\end{tcolorbox}
\caption{Formal language translation prompt without natural language alignment.}
\label{translation prompt no align}
\end{figure}

However, since the translator model is mainly trained on the dataset in the aligned format, we conduct an ablation study on whether natural language statement alignment is helpful or not. We use \texttt{meta-llama/Llama-3.3-70B-Instruct}\footnote{\href{https://huggingface.co/meta-llama/Llama-3.3-70B-Instruct}{https://huggingface.co/meta-llama/Llama-3.3-70B-Instruct}} as the alignment model to transform the original natural language statements into the corresponding aligned format. During translation, we keep the temperature as zero, and set the max length of new tokens to 1024. To accelerate the inference, we utilize the package vLLM \cite{kwon2023efficient} as our translation framework. The alignment prompt is shown as in Figure \ref{NL align prompt}.
\begin{figure*}[t]
\centering
\begin{tcolorbox}[title={Natural Language Statements Alignment Prompt}]
Example list:

====================\\
\`{}\`{}\`{} md\\
Let a,b,c>0. Prove that: \\
$\frac{a}{{b + c}}\sqrt {\frac{{bc}}{{\left( {b + a} \right)\left( {c + a} \right)}}} + \frac{b}{{c + a}}\sqrt {\frac{{ca}}{{\left( {c + b} \right)\left( {a + b} \right)}}} + \frac{c}{{a + b}}\sqrt {\frac{{ab}}{{\left( {a + c} \right)\left( {b + c} \right)}}} + \frac{{2abc}}{{\left( {a + b} \right)\left( {b + c} \right)\left( {c + a} \right)}} \leq 1$\\
\`{}\`{}\`{}\\
====================\\
\`{}\`{}\`{}md\\
Let $a,b,c \geq 0$ satisfy $a+b+c=3$ . Prove that \begin{align*}
(a^2+2bc)(b^2+2ca)(c^2+2ab) \leq\
\frac{81}{8}(ab+bc+ca)
\end{align*}\\
\`{}\`{}\`{}\\
====================\\
\`{}\`{}\`{}md\\
If $x_1+x_2=-a$ and $x_2+x_3=-b$ , then $x_1-x_3=b-a$\\
\`{}\`{}\`{}\\
====================\\
\`{}\`{}\`{}md\\
If $a,b,c \in[-k,k]$ where $a,b,c,k$ are real numbers and $a+b+c=0$ , show that $a^3+b^3+c^3 \le \frac{3k^3}{4}$\\
\`{}\`{}\`{}\\
====================\\
\`{}\`{}\`{}md\\
If $x,y,z>0 $ then prove: \\ $ x^2+y^2+z^2-xy-yz-zx+\frac{3xyz}{x+y+z}\ge\sqrt[3]{x^2y^2z^2} $\\
\`{}\`{}\`{}\\
====================\\
Help me transform the following math problem with the given answer to a proof problem, and the language format of the question should be identical to the example list above, also in the ```md``` structure. Do not modify the given answer even if it is not correct, just transform the format.
Question\\
\`{}\`{}\`{}md
\end{tcolorbox}
\caption{The prompt used for natural language statements alignment.}
\label{NL align prompt}
\end{figure*}


For a given math question answering problem in its original format, the alignment format will be in ``Given ... (premises), show that ... (goal)'', or ``If ... (premises), prove that ... (goal)'' like standard formats. Table \ref{table:nl align example} demonstrates a concrete example of aligning the original natural language statement into the standard format. From the results in table \ref{tab:compare trans nl}, we may notice that whether to use natural language alignment does not affect the overall performance of \fans, so we omit this alignment step in the other experiments.

\begin{figure*}[t]
\centering
\begin{tcolorbox}[title={Example of Natural Language Statements Alignment}]
        \textbf{Question} Convert the point $(0,3)$ in rectangular coordinates to polar coordinates.  Enter your answer in the form $(r,\theta),$ where $r > 0$ and $0 \le \theta < 2 \pi.$
        \tcbline
        \textbf{Answer} 
        Step 1: To convert from rectangular coordinates to polar coordinates, we use the formulas $r = \sqrt{x^2 + y^2}$ and $\theta = \arctan \left( \frac{y}{x} \right)$. Step 2: In this case, $x = 0$ and $y = 3$, so $r = \sqrt{0^2 + 3^2} = 3$ and $\theta = \arctan \left( \frac{3}{0} \right) = \frac{\pi}{2}$. Step 3: Since $0 \le \theta < 2 \pi$, we have $\theta = \frac{\pi}{2}$. Step 4: Therefore, the polar coordinates of the point $(0,3)$ are $\boxed{\left( 3, \frac{\pi}{2} \right)}$. The answer is: $(3,\frac{\pi}{2})$
        \tcbline
        \textbf{Aligned Output} Let $(x,y)=(0,3)$. Prove that the polar coordinates of the point are $(3,\frac{\pi}{2})$.
\end{tcolorbox}
    \caption{Example of the aligned natural language statement generated by the alignment model, given the original question and answer.}
    \label{table:nl align example}
\end{figure*}

The prompt used for formal language translation without Long-CoT is shown in Figure \ref{fl translate prompt no long-cot}, which takes advantage of the ability of LLMs' in-context learning provided few-shot examples.
\begin{figure*}[!h]
\centering
\begin{tcolorbox}[title={Formal Language Translation Prompt without Long-CoT}]
Convert the following natural language mathematical statement into a corresponding Lean4 statement.

\#\#\# Natural language statement:\\
Let a,b,c>0. Prove that:\\ 
$\frac{a}{{b + c}}\sqrt {\frac{{bc}}{{\left( {b + a} \right)\left( {c + a} \right)}}} + \frac{b}{{c + a}}\sqrt {\frac{{ca}}{{\left( {c + b} \right)\left( {a + b} \right)}}} + \frac{c}{{a + b}}\sqrt {\frac{{ab}}{{\left( {a + c} \right)\left( {b + c} \right)}}} + \frac{{2abc}}{{\left( {a + b} \right)\left( {b + c} \right)\left( {c + a} \right)}} \leq 1$

\#\#\# Lean4 statement:\\
\`{}\`{}\`{}lean4\\
theorem lean\_workbook\_plus\_49559 (a b c : $\mathbb R$) (ha : 0 < a) (hb : 0 < b) (hc : 0 < c) : (a / (b + c) * Real.sqrt (b * c / (b + a) * (c + a)) + b / (c + a) * Real.sqrt (c * a / (c + b) * (a + b)) + c / (a + b) * Real.sqrt (a * b / (a + c) * (b + c)) + 2 * a * b * c / (a + b) / (b + c) / (c + a)) $\le$ 1   :=  by 

\`{}\`{}\`{}\&

====================

Convert the following natural language mathematical statement into a corresponding Lean4 statement.

\#\#\# Natural language statement:\\
Let $a,b,c \geq 0$ satisfy $a+b+c=3$ . Prove that 
\begin{align*}
    (a^2+2bc)(b^2+2ca)(c^2+2ab)\leq\frac{81}{8}(ab+bc+ca)
\end{align*}

\#\#\# Lean4 statement:\\
\`{}\`{}\`{}lean4\\
theorem lean\_workbook\_plus\_82625 (a b c : $\mathbb R$) (ha : a $\ge$ 0 $\wedge$ b $\ge$ 0 $\wedge$ c $\ge$ 0) (hab : a + b + c = 3) : (a\^{}2 + 2 * b * c) * (b\^{}2 + 2 * c * a) * (c\^{}2 + 2 * a * b) $\le$ 81 / 8 * (a * b + b * c + c * a)   :=  by 

\`{}\`{}\`{}\&

====================

Convert the following natural language mathematical statement into a corresponding Lean4 statement.

\#\#\# Natural language statement:\\
\{natural language statement to be translated\}

\#\#\# Lean4 statement:\\
\`{}\`{}\`{}lean4
\end{tcolorbox}
\caption{Formal language translation prompt without Long-CoT instruction.}
\label{fl translate prompt no long-cot}
\end{figure*}

\begin{figure*}
    \begin{tcolorbox}[title={Formal Language Translation Prompt with Long-CoT}]
    Follow these instructions carefully:\\
1. Provide a logically correct and rigorous translate of natural language theorem statement to Lean4 statement.\\
2. In the <Thought> section, include your detailed step-by-step reasoning.\\
3. In the <Output> section, provide only the final Lean4 proof or final result.\\
\#\#\# Instruction:You will receive several Lean4 problems. For each:\\
- **Use** a step-by-step solution internally in <Thought>.\\
- **Do not** reveal your chain of thought outside the <Thought> block.\\
- **Ensure** the final Lean4 code or final result is placed **only** in <Output>.\\
@ Natural language theorem statement:\\
algebra\_sqineq\_2unitcircatblt1\\
Show that for any real numbers $a$ and $b$ such that $a^2 + b^2 = 2$, $ab \leq 1$.\\
@ Lean4 theorem statement:
\begin{lstlisting}
theorem algebra_sqineq_2unitcircatblt1 (a b : ℝ) (h₀ : a ^ 2 + b ^ 2 = 2) : a * b ≤ 1 :=
\end{lstlisting}
====================\\
@ Natural language theorem statement:\\
mathd\_numbertheory\_629\\
Suppose $t$ is a positive integer such that $\mathop{\text{lcm}}[12,t]^3=(12t)^2$. What is the smallest possible value for $t$? Show that it is 18.\\
@ Lean4 theorem statement:
\begin{lstlisting}
theorem mathd_numbertheory_629 : IsLeast { t : ℕ | 0 < t ∧ Nat.lcm 12 t ^ 3 = (12 * t) ^ 2 } 18 :=
\end{lstlisting}
====================\\
@ Natural language theorem statement:\\
\{title\} \{NL statement\}\\
@ Lean4 theorem statement:\\
\#\#\# Response:\\
<Thought>\\
Alright, I should do the following:\\
1. Analyze the problem and understand the requirements.\\
2. Use natural language to analyze the potential Lean4 condition and requirements in the natural language theorem statement.\\
3. Write the translated Lean4 statement code.\\
The \`{}\{title\}\`{} has the following conditions
    \end{tcolorbox}
    \caption{Formal Language Translation Prompt with Long-CoT}
    \label{fl translate prompt long-cot}
\end{figure*}


\begin{figure*}[!h]
\begin{tcolorbox}[title={Header of Formal Proof Prompt}]
Complete the following Lean 4 code with explanatory comments preceding each line of code:

\`{}\`{}\`{}lean4\\
import Mathlib\\
import Aesop\\

set\_option maxHeartbeats 0

open BigOperators Real Nat Topology Rat
\end{tcolorbox}
\caption{Header of Formal Proof Prompt}
\label{fl proof prompt header}
\end{figure*}

Concatenating the formal language statement and the proof header, together with the natural language statement as the auxiliary comment, we trigger the prover model and generate potential proofs for the given formal language statements.

Below we present more concrete examples from \fans, verifying that with the assistance of formal language, we could select the correct answer out of all possible candidates successfully.
\begin{figure*}[t]
\centering
\begin{tcolorbox}[title={More Concrete Examples for \fans Pipeline}]
        \textbf{Natural Language Math Problem}\\ Find the value of $\log_2{3} \cdot \log_3{4} \cdot \log_4{5} \cdot \log_5{6} \cdot \log_6{7} \cdot \log_7{8}$.\\
        \textbf{Answer} 3\\
        \textbf{Translated Formal Language Statement} 
        \begin{lstlisting}
theorem math_problem_0 :  Real.logb 2 3 * Real.logb 3 4 * Real.logb 4 5 * Real.logb 5 6 * Real.logb 6 7 * Real.logb 7 8 = 3 := by
        \end{lstlisting}
        \textbf{Formal Language Proof}
        \begin{lstlisting}
theorem math_problem_0 : Real.logb 2 3 * Real.logb 3 4 * Real.logb 4 5 * Real.logb 5 6 * Real.logb 6 7 * Real.logb 7 8 = 3 := by
  simp [Real.logb, mul_assoc, mul_comm, mul_left_comm, show (3 : ℝ) = 3 by norm_num,
    show (4 : ℝ) = 2 ^ 2 by norm_num, show (5 : ℝ) = 5 by norm_num, show (6 : ℝ) = 2 * 3 by norm_num,
    show (7 : ℝ) = 7 by norm_num, show (8 : ℝ) = 2 ^ 3 by norm_num]
  field_simp
  ring_nf
        \end{lstlisting}
        \vspace{-1.5em}
    \tcbline
    \textbf{Natural Language Math Problem}\\
    A steel sphere with a 3-inch radius is made by removing metal from the corners of a cube that has the shortest possible side lengths. How many cubic inches are in the volume of the cube?\\
    \textbf{Answer} 216\\
    \textbf{Translated Formal Language Statement}
    \begin{lstlisting}
theorem math_problem_0 (r : ℝ) (h₀ : r = 3) : let cube_side := 2 * r; cube_side ^ 3 = 216 := by
    \end{lstlisting}
    \textbf{Formal Language Proof}
\begin{lstlisting}
theorem math_problem_0 (r : ℝ) (h₀ : r = 3) : let cube_side := 2 * r; cube_side ^ 3 = 216 := by
  subst h₀
  simp only [pow_three]
  norm_num
\end{lstlisting}
\end{tcolorbox}
    \caption{More concrete examples that demonstrate formal language could indeed translate the natural language statement correctly and then prove it successfully.}
    \label{more fans examples}
\end{figure*}

\end{document}